# 小型无人直升机 MIMO 系统飞行鲁棒控制

何淼磊[1],贺继林[1]

(中南大学 机电工程学院,长沙 410083)

针对小型无人直升机的全自主飞行控制问题,提出了一种基于 H∞鲁棒控制器的设计方法,设计双闭环的飞行控制器。内环控制器采用 H∞鲁棒控制技术用于直升机飞行姿态的控制,外环控制器采用 PD 控制技术用于机体位移的控制,将设计的鲁棒控制器应用于研制的小型直升机飞行控制系统,实现了风扰环境下直升机的全自主定点悬停飞行。实验验证了该控制方法的有效性和鲁棒性。

关键字:无人直升机;飞行控制;鲁棒控制

## Robust Flight Control of Small Unmanned Helicopter MIMO System

HE Miaolei[1], HE Jilin[1]

(College of Mechanical and Electrical Engineering, Central South University, Changsha 410083)

Abstract: For small unmanned helicopter autonomous flying control problem, a flight controller design method of robust H∞ controller using double closed-loop is presented. The inner-loop controller uses H∞ robust control technology for helicopter flight attitude controlling, the outer-loop controller uses PD control technology for control the helicopter displacement controlling, the robust controller is applied to flight control system of the Independently developed helicopter, realize the fully autonomous fixed-point hovering flight under interference environment. Experiment demonstrate the effectiveness and robustness of this control approach.

## 1 引言(Introduction)

小型无人直升机具有高机动性、小尺寸、低成本的特点,在工业、农业、军事等领域具有广泛的用途。无人直升机由于其自身复杂的动力学特性,以及直升机系统非线性、多变量输入输出、强耦合的特点,飞行控制系统的设计成为一个研究热点[1-5]。对于无人直升机飞行控制的研究,国内外有很多科研机构了进行了相关研究工作,在准确获得直升机非线性模型基础上,有些还开发了无人直升机试验平台[6]。在小型无人直升机的控制器实现上,经典的控制的方法有 PID 控制[7-8]、μ 综合飞行控制器[9]、神经网络控制器[10]、滑模控制器[11]和反步控制[12]等。

本课题组研制的小型无人直升机控制系统基于航模汽油直升机本体,自主研制开发了飞行控制系统。机载飞行控制器以 TMS320F2812 DSP 芯片为飞控处理器,导航部分包括惯性测量单元、驱动输出模块和 GPS 模块等,飞行数据采集系统通过机载存储卡保存。

本文提出了基于 H∞鲁棒控制技术的小型直升机控制器设计方法,设计了小型无人直升机的双闭环结构的飞行控制器并将其用于无人直升机的全自主飞行,实现了稳定的定点悬停。最后,通过在无人直升机上的实验,验证了该控制方法的有效性。

## 2 直升机动力学模型(Helicopter dynamic model)

采用先进的控制技术设计直升机飞行控制系统,必须获得能够反映直升机飞行动力学特性的数学模型,直升机的非线性模型主要包括四个部分:机体运动学特性、机体动力学特性、主旋翼挥舞动力学特性和偏航角速率控制器动态特性。

### 2.1 机体运动学特性

机体运动学特性包括两部分平移运动和旋转运动,假设本地 NED(north-east-down)轴系是惯性系,且机载 NED 坐标系与之方向相同。

对于直升机机体平移运动,公式如下:

$$h = -P_{ned\_z}$$
$$[\dot{P}_{ned\_x}, \dot{P}_{ned\_y}, \dot{P}_{ned\_z}]^T = R_B[V_x, V_y, V_z]^T \qquad (1)$$

其中 $P_{ned\_x}$, $P_{ned\_y}$, $P_{ned\_z}$ 表示直升机在本地 NED 三维坐标系中的位移,$h$ 表示直升机飞行高度。$V_x$,$V_y$,$V_z$ 表示机体坐标系中直升机对地相对速度。$R_B$ 表示从机载 NED 轴系到机体轴系的旋转矩阵,公式如下:

$$R_B=\begin{bmatrix} C_\theta C_\phi & C_\phi C_\psi & -S_\phi \\ S_\phi S_\theta C_\psi - C_\phi C_\psi & S_\phi S_\theta S_\psi + C_\theta C_\psi & C_\phi S_\theta \\ S_\phi C_\theta S_\psi + S_\theta S_\psi & S_\phi C_\theta C_\psi - S_\theta C_\psi & C_\phi C_\theta \end{bmatrix} \quad (2)$$

式中，$C^*$、$T^*$和$S^*$分别表示余弦、正切与正弦函数。

对于直升机机体的旋转运动，公式如下：

$$[\dot\phi,\dot\theta,\dot\psi]^T = s_b \omega_{b/n}^b \quad (3)$$

式中$\omega_{b/n}^b$表示机体轴系角速度向量，式中$s_b$由下式给出：

$$s_b = \begin{bmatrix} 1 & T_\theta S_\phi & T_\theta C_\phi \\ 0 & C_\phi & -S_\phi \\ 0 & C_\phi/S_\theta & C_\varphi/C_\theta \end{bmatrix} \quad (4)$$

## 2.2 刚体动力学特性

根据 Newton-Euler 方程以及上文关于本地 NED 的假设，可以导出直升机机体的六自由度刚体动力学特性：

$$\begin{aligned} \dot V_b &= -[\varphi,\theta,\psi]^T \times V_b + \frac{F_b}{m} + \frac{F_{b.g}}{m} \\ \dot\omega_{b/n}^b &= J^{-1}\left[M_b - \omega_{b/n}^b \times \left(J\omega_{\frac{b}{n}}^b\right)\right] \\ J &= \mathrm{diag}(J_X, J_Y, J_Z) \end{aligned} \quad (5)$$

式中 $m$ 表示直升机质量，$J$ 是惯性矩矩阵，$F_b$ 是气动力矢量，$F_{b.g}$ 是直升机重力矢量在机体坐标系上的投影，$M_b$ 是气动力矩矢量。对于小型直升机，惯性矩矩阵主对角线之外的元素影响非常小，建过程中可以忽略。

## 2.3 主旋翼动态特性

主旋翼桨尖轨迹平面的纵向与横向挥舞角分别定义为$a_s$，$b_s$，从主桨周期变矩到其挥舞角的挥舞动力学可以用两个耦合的一阶微分方程描述：

$$\begin{aligned} \dot a_s &= -q - \frac{1}{\tau_{mr}} a_s + A_{b_s} + \frac{1}{\tau_{mr}} \theta_{a_s} \\ \dot b_s &= -p - \frac{1}{\tau_{mr}} b_s + B_{b_s} + \frac{1}{\tau_{mr}} \theta_{b_s} \end{aligned} \quad (6)$$

式中$\tau_{mr}$表示主旋翼挥舞运动时间系数，$\theta_{a_s}$和$\theta_{b_s}$表示桨叶纵向桨距角和横向桨距角。$A_{b_s}$和$B_{b_s}$表示纵向挥舞与横向挥舞运动的耦合系数，其表达式如下：

$$A_{b_s} = -B_{b_s} = \frac{8k_\beta}{\gamma_{mr}\Omega_{mr}^2 I_\beta} \quad (7)$$

## 2.4 偏航角速率反馈控制器

无人直升机偏航控制对操作极其敏感，操作手难以操作，目前几乎所有的小型无人直升机都安装了航向锁定装置。为了便于实现手动操作切换到自主飞行状态，直升机平台保留了该部分，因此，在模型中需要引入偏航角速率反馈控制器。偏航角速率反馈控制器中，控制信号输入$\delta_{ped}$通过比例放大电路放大信号，然后与角速率陀螺仪检测到的反馈值 $r$ 比较，偏差信号送入控制器，生成尾桨舵机偏转信号$\delta_{ped}'$，控制策略采用经典的 PI 控制，动态特性如下：

$$\delta_{ped}' = \left(K_P + \frac{K_I}{s}\right)(K_a \delta_{ped} - r) \quad (8)$$

## 3 控制器设计（Controller design）

无人直升机双闭环飞行控制框图如下图所示，其中内环用于直升机欧拉角$\phi$、$\theta$、$\psi$，角速度 $p$、$q$、$r$ 以及偏航角速度控制器内部中间状态相关的飞行器动力学特性；控制器的外环主要调节直升机的机体轴系的空速矢量以及本地坐标系的位置$P = (P_{ned\_x}, P_{ned\_y}, P_{ned\_z})$。由于位置信息是由 GPS 数据与速度积分后的数据融合后给出，因而动态特性较内环控制回路低，这种控制回路结构对于无人直升机系统来说是比较适合的。

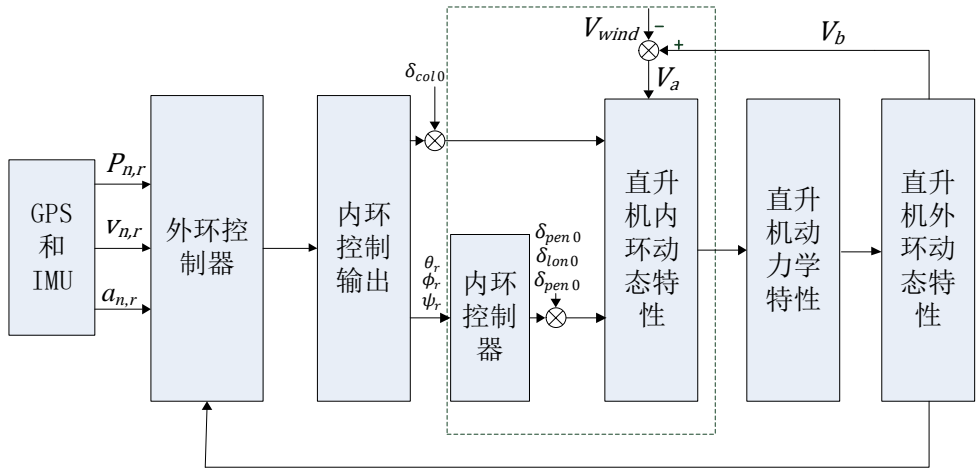

图 1 无人直升机双闭环飞行控制框图

Fig.1 Structure of unmanned helicopter dual closed-loop control system

### 3.1 内环控制器设计

直升机内环动态特性高，直接关系到直升机机体的稳定性，更直接影响外环控制器的控制品质，是整个直升机控制系统的关键。在内环控制器的设计过程中，引入环境干扰因素中的风速，采用 H∞ 最优技术抑制干扰，将风速干扰的影响降到最小。

在直升机平衡点附近将上述非线性模型线性化，动态模型的线性化模型如下：

$$\dot{x} = Ax + Bu + EV_{wind}$$
$$y = [\phi, \theta, p, q, r, \psi] - y_{trim} \tag{9}$$

式中 x=$x_{act} - x_{trim}$ 是实际状态变量与相应配平值之差，类似的，u=$u_{act} - u_{trim}$。$V_{wind} = [u_{wind}, v_{wind}, w_{wind}]^T$ 表示风速沿机体轴系的风速分量，矩阵 E 可通过在对应通道注入阵风干扰的实验获得。$x_{act}$ 和 $u_{act}$ 分别如下：

$$x_{act} = [\phi, \theta, p, q, a_s, b_s, r, \delta_{ped}', \psi]$$
$$u_{act} = [\delta_{lat}, \delta_{lon}, \delta_{ped}]^T \tag{10}$$

上式中，$\phi$ 为滚转角，$\theta$ 为俯仰角，$p$ 为滚转角速度，$q$ 为俯仰角速度，$a_s$ 为横向挥舞角，$b_s$ 为纵向挥舞角，r 为偏航角速度，$\delta_{ped}'$ 为偏航角速度反馈控制器的内部状态变量，$\psi$ 为偏航角；$\delta_{lat}$ 为滚转舵机输入，$\delta_{lon}$ 为俯仰舵机输入，$\delta_{ped}$ 为偏航角速度反馈控制控制器输入，舵机输入的取值范围为(-1，1)。

内环控制器的任务主要是负责镇定直升机机体姿态角的变化，因此控制器的设计中并没有使用总距输入$\delta_{col}$。为了保证系统具有较好的姿态相应，因此基本被控输出变量选择为 $h_{out}$=$[\phi,\theta,\psi]^T$-$h_{out,trim}$，式中 $h_{out,trim}$ 表示 $h_{out}$ 在实验中测得的配平值。

内环控制器设计中，为了约束输入变量其他状态变量，采用以下被控输出：

$$h = Cx + Du \tag{11}$$

式中 $C$ 和 $D$ 为常数矩阵：

$$C = \begin{bmatrix} 0_{3\times 9} \\ C_{11} & 0_{4\times 5} \\ 0_{2\times 4} & C_{22} \end{bmatrix}$$
$$D = \begin{bmatrix} D_{11} \\ 0_{6\times 3} \end{bmatrix} \tag{12}$$

$C_{11}$、$C_{22}$ 和 $D_{11}$ 是待定的权重系数矩阵。H∞ 输出控制率的设计也就是风速干扰 $V_{wind}$ 到被控输出变量 $h_{out}$ 的 H∞ 范数，其传递函数的 H∞ 范数定义为：

$$\|T_{hv}\|_\infty := \sup \sigma_{max}[T(jw)] = \sup \frac{\|h\|_2}{\|V_{wind}\|_2} \tag{13}$$

式中$w \in [0, \infty)$，$\|V_{wind}\|_2 = 1$。很显然，传递函数矩阵的 H∞范数对应输入到输出在最差情况下的增益，为了控制器设计方便，定义：

$$\gamma^* := \inf \{\|T_{hv}(G \times F)\|_\infty\} \quad (14)$$

$G$ 为内环控制器中前向矩阵，$F$ 为反馈增益矩阵，系统前向矩阵以及反馈增益矩阵如下：

$$G = -[C'(A + BF)^{-1}B]^{-1}$$
$$F = -(D^TD)^{-1}(D^TC + B^TP) \quad (15)$$

控制器的设计中，$\gamma^*$值的确定相当繁琐,通常采用迭代的方法,通过多次迭代运算虽然可以获得精度满意的$\gamma^*$，但是运算过程代价很高，寻优过程繁琐，更为重要的是在接近$\gamma^*$时，可能产生较高的增益，使得求的结果不具有鲁棒性甚至具有很高的病态。所以采用计算带参数的$\gamma$的 Riccati-Equation 解来迭代求取$\gamma^*$稳定性差，这里我采用文献[16]的方法精确计算$\gamma^*$。对于 H∞控制器来说，很难获得最优性能，因此只能设计 H∞的次优控制律，实际中选定的标量为$\gamma = 0.0632$。

P 为 H∞代数 Riccati-Equation 的半正定解：

$$PA + A^TP + C^TC + PEE^TP/\gamma^2 - (PB + C^TD)(D^TD)^{-1}(D^TC + B^TP) = 0 \quad (16)$$

同时，很容易验证矩阵 D 为列满秩，矩阵（A,B,C,D）左可逆同时没有不变零点，设计的 H∞的$\gamma$次优状态反馈控制律如下：

$$u = Fx + G(r - h_{out,trim}) \quad (17)$$

在内环控制器的设计中，需要通过地面实验对某些无法直接测量的物理量进行实验获得。直升机悬停（速度小于 5 米/秒）是直升机飞行模态中较为典型的飞行模态，其他飞行模态多为经验形式的定义，难以根据数据准确判定，同时悬停模态下的直升机工作点范围宽，数据采集危险性最低。经过多次实验，获得被控输出中的权重系数矩阵$C_{11}$、$C_{22}$和$D_{11}$。

$$C_{11} = \begin{bmatrix} 13 & 0 & 0 & 0 \\ 0 & 11 & 0 & 0 \\ 0 & 0 & 1 & 0 \\ 0 & 0 & 0 & 1 \end{bmatrix} \quad C_{22} = \begin{bmatrix} 0 & 0 & 1 & 0 & 0 \\ 0 & 0 & 0 & 0 & 5 \end{bmatrix}$$

$$D_{11} = \begin{bmatrix} 12 & 0 & 0 \\ 0 & 11 & 0 \\ 0 & 0 & 31 \end{bmatrix} \quad (18)$$

悬停状态下状态变量和输入变量的配平值见表1，未列出的变量配平值取0。

表 1 悬停状态下无人直升机变量配平值

Tab.1 Unmanned helicopter variable value in hovering

| 变量名称 | 配平值 | 变量名称 | 配平值 |
| --- | --- | --- | --- |
| $\phi$/(rad) | 0.0287 | $\delta_{lat}$/(NA) | 0.0075 |
| $\theta$/(rad) | 0.0011 | $\delta_{lon}$/(NA) | -0.0060 |
| $a_s$/(rad) | -0.0011 | $\delta_{col}$/(NA) | -0.1793 |
| $b_s$/(rad) | 0.0042 | | |

$x_{act}$中 $a_s$、$b_s$ 和$\delta_{ped}'$三个变量很难直接测量，这里需要设计一个降阶观测器，降阶观测器如下：

$$\begin{bmatrix} a_s \\ b_s \\ \delta_{ped}' \end{bmatrix} = x' + Ky$$
$$\dot{x}' = A'x' + B'y + H'u \quad (19)$$

用式（19）中不可测量变量的表达式代替反馈控制律（17）中变量，矩阵$A'$、$B'$、$H'$、$K$矩阵的参数，同样采用地面实验结合参数辨识的方法获得。

### 3.2 外环控制器设计

无人直升机外环控制器主要控制直升机位置矩阵，其动态特性较内环低，内环设计的鲁棒控制器可以精确的控制直升机的姿态角，外环控制器的控制性能是受内环控制器的控制效果影响的，因此实现内环精确的直升机姿态控制，可以简化外环控制的设计过程，同时又能保证理想的控制效果。外环控制器采用 PD 控制[4]，设计的高度控制

器如下：

$$P_{ned\_z} = \frac{1}{C_\phi C_\theta}(K_{P,ned\_z}(P'_{ned\_z} - P_{ned\_z}) + K_{D,ned\_z}(\dot{P}'_{ned\_z} - \dot{P}_{ned\_z}) + mg) \quad (20)$$

中$P'_{ned\_z}$、$P'_{ned\_z}$表示高度的参考值，$1/(C_\phi C_\theta)$是直升机运动升力补偿系数。X、Y 轴的位置控制器设计类似，这里以 X 轴为例。设计的 X 轴位移控制器如下：

$$\theta = \arcsin(K_{P,ned_x}(P'_{ned\_x} - P_{ned\_x}) + K_{D,ned\_z}(\dot{P}'_{ned\_x} - \dot{P}_{ned\_x})) \quad (21)$$

内环姿态控制器的输入作为外环上述控制器的输出量。

## 4 实验验证（Experiment validation）

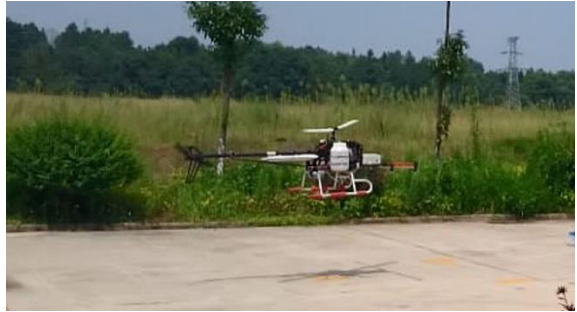

图 2 全自主定点悬停飞行中的无人直升机

Fig.2 Unmanned helicopter in automatic spot hovering

将设计的鲁棒控制器用于无人直升机实验平台的自主飞行控制，测试地点选择较为空旷的场地，有利于机载 GPS 的搜星定位。采用 GPS 和惯性器件组合导航的策略， GPS 刷新率较低造成的位置精度偏差较大的问题，进行了在全自主定点悬停飞行实验，并将鲁棒控制器控制的直升机姿态实验结果与 PID[7]控制的实验结果进行比较。

飞行姿态控制的实验中，环境风速≈3m/s，操作直升机启动后，通过模式切换到指令飞行状态，即横滚、俯仰角给定为 0°，直升机姿态数据通过机载的数据卡记录，采集的横滚角和俯仰角姿态数据如图所示。出 PID 控制器控制的直升机姿态角精度范围±4.8°，而鲁棒控制器可将直升机姿态角精度控制在±1.7°。从图 4 中可以看出采用鲁棒控制器控制的直升机飞行时，环境风速的干扰得到了很好的抑制，姿态角波动小。

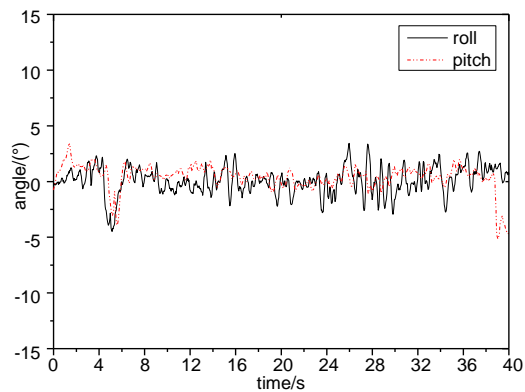

图 3 基于 PID 控制的横滚、俯仰角响应曲线

Fig.3 Response of roll and pitch under PID control

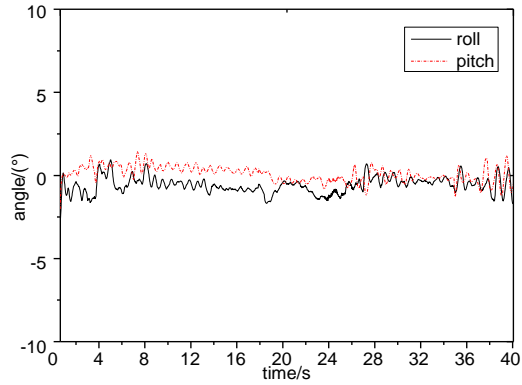

图 4 基于鲁棒控制的横滚、俯仰角响应曲线

Fig.4 Response of roll and pitch under robust control

在定点悬停实验过程中，操作直升机启动后，带直升机到达悬停点时将飞行模式切换到自主定点悬停的飞行模态，实验中地面风速≈3m/s，此时所设计的控制器内环、外环控制器均参与控制指令输出。在低空定点悬停飞行 18s 后，直升机以 2.5m/s 的速度上升，22s 时刻停止上升，开始定点悬停飞行，在附近等高区域测得的风速约 5m/s。直升机体垂向的运动速率如图 5 所示，Z 轴的速率控制精度达到±0.18m/s。机体 X、Y 轴的悬停速率如图 6、图 7 所示，从实验结果可以看出，X、Y 轴速率控制精度达到±0.3m/s。从相应曲线的对比还可以看出：Z 轴方向的运动是自然解耦的，系统的动态性好；X、Y 轴上运动的耦合效应明显，系统的惯性导致控制输出具有一定的滞后性。

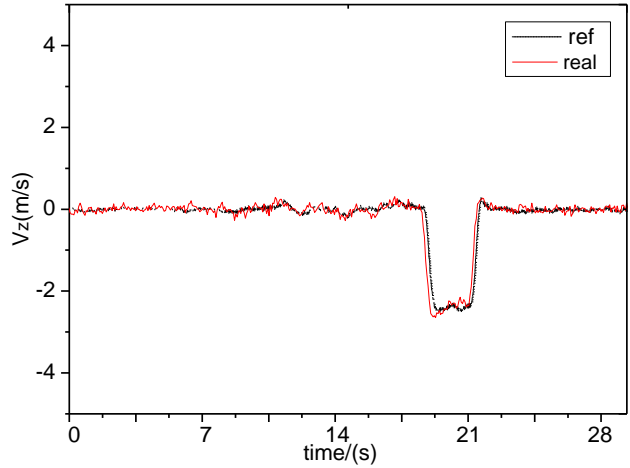

图 5 Z 轴速率响应曲线

Fig.5 Response of velocity along Z

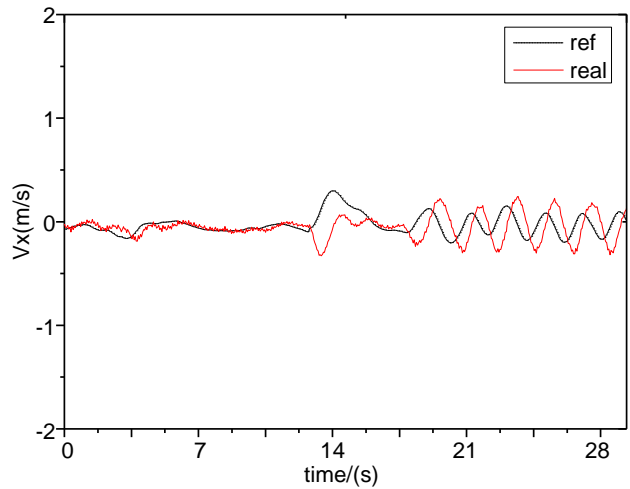

图 6 X 轴速率响应曲线

Fig.6 Response of velocity along X

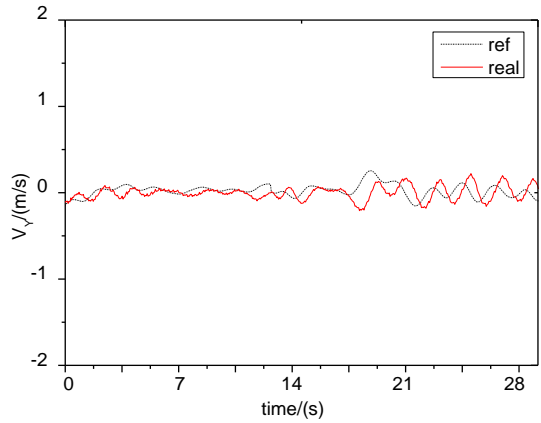

图 7 Y 轴速率响应曲线

Fig.7 Response of velocity along Y

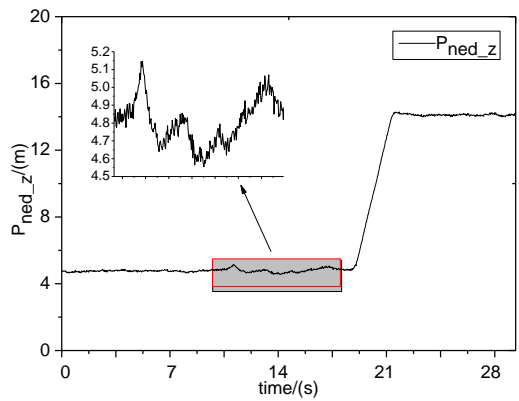

图 8 悬停飞行中高度曲线

Fig.8 Response of velocity along Y

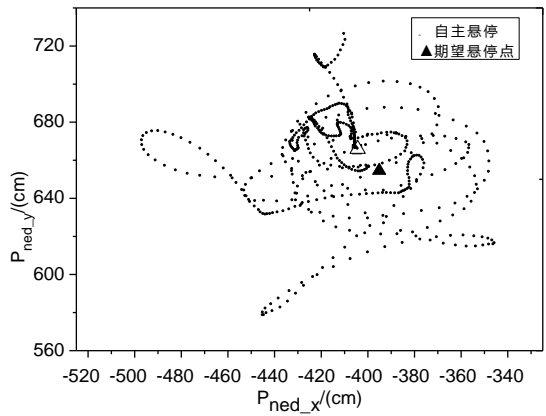

图 9 水平位置轨迹曲线

Fig.9 Horizontal position trajectory

图 8 给出了定点悬停飞行时机体垂向运动的高度曲线图，图 9 给出了定点悬停飞行时机体水平运动轨迹图。图 8 中将高度波动较大的时间段放大后可以看出，垂向的飞行高度控制精度小于 40cm，直升机飞行高度上升后，高空风速大于地面风速，干扰加强，如图 6、图 7 的速率波动曲线所示。从图 9 水平运动轨迹的实验数据来看，在整个飞行过程中，机体横向位移控制的位置误差小于 102cm，纵向位移控制的位置误差小于 65cm。类似的实验中，澳大利亚的无人直升机采用 PD-PID 控制方法[8] 设计的直升机室外悬停位置偏移误差小于 1.2 米。新加坡国立大学采用非线性控制器设计的直升机室外悬停位置偏移误差小于 1.33 米[17]。通过比较可知，本文的控制器的设计方法是有效的，同时实验结果验证了上节中关于外环控制器的设计思路。

## 5 结论（Conclusion）

本文针对小型无人直升机的控制问题，提出了一种基于 H∞控制的鲁棒控制器设计方法，将设计的双闭环鲁棒控制器应用于实验直升机。实验结果表明，设计的控制器实现了对直升机姿态角、位移的精确控制，系统具有较好的鲁棒性和抗干扰能力，为小型无人直升机的在阵风环境下的定点悬停控制提供了一种新的思路。

接下来的工作中，尝试采用其他控制策略，以实现无人直升机在大机动飞行模态下的飞行姿态和飞行轨迹的精确控制。

## 参考文献（References）

**作者简介：**

何淼磊（1987--），男，博士生，研究领域：无人机飞行系统。

贺继林（出生年--），男，博士，副教授，博士生导师，研究领域：航空装备与特种机器人、机电液一体化。